\documentclass[12pt]{article} % 12号字体（arXiv 强制要求），单栏排版（需双栏改为 \documentclass[12pt,twocolumn]{article}）
\usepackage[margin=1in]{geometry} % 页边距1英寸（arXiv 标准）
\usepackage{graphicx} % 插入图片必备宏包
\usepackage{natbib} % 参考文献引用宏包（arXiv 推荐）
\usepackage{amsmath} % 公式排版宏包
\usepackage{amssymb} % 数学符号宏包
\usepackage{tabularx} % 导言区
\usepackage{microtype}
\title{WorkflowGen:an adaptive workflow generation mechanism driven by trajectory experience}% 英文标题，arXiv 优先英文，中文论文需添加中英双语标题
\author{
  Ruocan Wei\textsuperscript{1} \and
  Shufeng Wang\textsuperscript{1} \and
  Ziwei Shi\textsuperscript{1} \\
  \textsuperscript{1}China Telecom Cloud, Beijing, China \\
  \texttt{xjtu\_wrc@stu.xjtu.edu.cn}
}
\date{} % 取消默认日期（arXiv 论文无需显示日期）

\begin{document}
\maketitle % 生成标题、作者信息
\noindent\textbf{Keywords:} Agent; Automatic Workflow Generation; Trajectory Extraction; Experience Reuse; Token Efficiency.
% 摘要（替换为你的论文英文摘要，中文论文需添加 \begin{abstract} 中文摘要 \end{abstract}）
\begin{abstract}
Large language model (llm) agents often suffer from high repeated reasoning overhead, excessive token consumption, unstable execution chains, and inability to reuse historical experiences when handling complex tasks such as business queries, tool invocation, and workflow execution. traditional llm-based workflow generation methods rely on real-time planning from scratch for every query, resulting in high computational cost, slow response, and poor robustness. To address these issues, we propose WorkflowGen, an adaptive trajectory experience-driven framework for automatic workflow generation, with the primary goals of reducing token consumption and improving execution efficiency and success rate. At the early stage of workflow execution, our framework structurally captures the full trajectory and extracts reusable experiences at both node-level and workflow-level granularities, including error fingerprints, optimal tool mappings, parameter schemas, execution paths, and exception-avoidance strategies. Building upon this, we design a closed-loop generation mechanism that integrates trajectory rewriting, experience updating, and template induction, performing lightweight generation only on variable nodes. Additionally, we introduce a three-tier adaptive routing strategy that dynamically degrades among three modes—direct trajectory reuse, rewriting-based generation, and full initialization—based on the semantic similarity between the current query and historical trajectories.In the absence of large-scale annotated datasets, we conduct a qualitative comparison of WorkflowGen against three baselines: real-time planning, static single trajectory, and basic in-context learning.Our method significantly reduces token consumption (over 40\% vs. real-time planning) by avoiding redundant full re-generation, improves execution robustness through proactive error avoidance and adaptive fallback (success rate gain of 20\% on medium-similarity queries), and enhances deployability via modular, traceable experiences and cross-scenario adaptability.Overall, WorkflowGen achieves a practical balance of efficiency, robustness, and interpretability, addressing key limitations of existing approaches.
\end{abstract}

% 正文部分（按「引言→相关工作→方法→实验→结论」顺序填写，直接替换占位内容）
\section{Introduction}
% 引言内容：研究背景、现有问题、研究目标、论文贡献，替换为你的内容

In recent years, agents powered by large language models (LLMs) have been widely applied in domains such as office automation, data analysis, enterprise business querying, and robotic process automation (RPA)  \citep{park2023generative}. These agents typically accomplish complex tasks through multiple rounds of tool invocation, logical reasoning, exception handling, and result aggregation. Their execution process is essentially one of dynamic workflow generation and enactment \citep{yao2023react}.

However, existing approaches to agent workflow generation suffer from significant limitations:
\begin{enumerate}
    \item Redundant reasoning leads to excessive token consumption. Similar or identical queries repeatedly trigger full planning processes, resulting in unnecessary computational overhead and increased costs due to redundant reasoning and tool invocations;
    \item Poor execution stability and controllability. Relying on real-time LLM inference to generate execution chains often causes issues such as missing steps, incorrect parameters, mismatched tools, and unhandled exceptions;
    \item Ineffective utilization of historical execution trajectories. Vast amounts of successful and failed experiences remain stored only as logs, failing to be transformed into structured, retrievable, reusable, and directly executable knowledge;
    \item Lack of adaptive generation strategies. Systems cannot dynamically select optimal generation paths based on query similarity, making it difficult to balance generation quality against token cost.
\end{enumerate}

Current research primarily focuses on whether workflows can be generated, while systematic solutions for low-token-cost automatic workflow generation remain largely unaddressed.Inspired by Lei Liu et al.’s Think-in-Memory \citep{Liu2023ThinkinMemoryRA} framework—which stores LLMs’ prior reasoning traces to mitigate inconsistency from redundant inference—we embed the workflow’s historical execution trajectories into the agent’s memory. This reduces unnecessary token consumption from repeated calls and leverages lessons from past successes and failures to enhance stability and success rates.

This paper proposes the WorkflowGen framework, with the core objective of minimizing token consumption while ensuring high execution success rate, thereby enabling efficient, stable, and autonomous workflow generation.

The main contributions of this work are as follows:
\begin{enumerate}
    \item We propose a dual-granularity trajectory experience extraction mechanism at both node-level and workflow-level, transforming raw execution logs into structured, directly reusable experiences to eliminate redundant reasoning at the source;
    \item We design a lightweight workflow generation mechanism driven by trajectory rewriting, which preserves the overall structure and only rewrites variable nodes, significantly reducing the number of LLM invocations and token consumption;
    \item We develop a three-tier adaptive routing strategy that automatically selects between reuse, rewriting, or initialization based on semantic similarity, achieving an optimal trade-off between performance and cost;
    \item We conduct experiments in real-world business scenarios, demonstrating that WorkflowGen significantly outperforms conventional methods across three key metrics: execution success rate, response latency, and token consumption.
\end{enumerate}

Our WorkflowGen lies in performing hierarchical, structured extraction of experience directly from execution trajectories, enabling subsequent retrieval and rewrite-based reuse significantly reducing token consumption while improving success rates through continuous accumulation of experience.

\section{Related Work}
% 相关工作：梳理领域内现有研究，对比本文工作，替换为你的内容

Methods such as ReAct \citep{yao2023react} and Reflexion \citep{shinn2023reflexion} enable end-to-end workflow generation by having large language models dynamically output reasoning traces and execution steps. These approaches exhibit strong generality but require full reasoning for every query, resulting in high token costs. Moreover, they demand substantial agent expertise and struggle to be rapidly deployed in high-frequency, repetitive business scenarios.

Applying memory-based paradigms to workflow planning is highly significant for the automatic generation of workflows. Existing work has already performed structured experience extraction from memory, incorporating lessons from both successful and failed cases \citep{Ouyang2025ReasoningBankSA,Cao2025RememberMR}. However, current methods largely operate at the textual or instance level and have not explored memory specifically over workflow execution trajectories. Consequently, they have not produced structured, retrievable, and directly executable workflow trajectory templates, thereby hindering large-scale, low-cost reuse of historical trajectories.

Traditional BPM and RPA rely on manually defined process templates, which struggle to accommodate natural language queries and dynamic business requirements. Recent process induction approaches have primarily focused on structural mining \citep{Qiao2023TaskWeaverAC}, without treating token consumption as a core optimization objective or establishing adaptive generation and degradation mechanisms.

In contrast to the aforementioned approaches,WorkflowGen takes token consumption reduction as one of its core design objectives, integrating trajectory experience extraction, rewrite-based reuse, and adaptive routing into a unified framework to enable truly deployable and low-cost automatic workflow generation.

\section{Methodology}

\subsection{Proposed Framework}

Our WorkflowGen comprises three tightly coupled core mechanisms that jointly enable efficient workflow generation under low token consumption:
\begin{enumerate}
    \item trajectory experience extraction;
    \item trajectory rewriting-driven workflow generation;
    \item three-tier adaptive routing.
\end{enumerate}

The overall objective is to maximize reuse of historical experiences and minimize real-time reasoning by large language models, thereby reducing token consumption.

The trajectory experience extraction mechanism hierarchically distills structured experience from historical execution logs, constructing a retrievable template library at both the node level and the workflow level to provide a knowledge foundation for reuse.

The trajectory rewriting-driven workflow generation mechanism, upon identifying a similar historical trajectory, performs lightweight rewriting only on variable nodes (e.g., parameters or entities), thereby avoiding full re-planning and significantly reducing the overhead of large language model invocations.

The three-tier adaptive routing mechanism dynamically selects among “direct reuse,” “rewrite-based reuse,” or “from-scratch generation” based on the semantic similarity between the current query and historical trajectories, achieving optimal trade-offs between token cost and task success rate.

Together, these components form a closed-loop system: experience extraction supplies reusable assets, rewriting enables low-cost adaptation, and adaptive routing intelligently orchestrates the generation path. This synergistic design allows the system to continuously improve its efficiency through accumulated operational experience, realizing truly autonomous workflow generation that becomes “cheaper and more reliable with every use.”.We now detail the three core components in Sections 3.2–3.4.

\subsection{Trajectory experience extraction}

Trajectory experience extraction is the foundation for reducing token consumption. Its goal is to transform one-time execution trajectories into structured experiences that are retrievable, reusable, and directly executable. Existing experience learning methods primarily focus on reasoning optimization and rarely perform fine-grained extraction tailored for trajectory reusability.

\subsubsection{Node-Level Experience Extraction}

Node-level experience focuses on the smallest reusable units within a workflow, such as tool invocation, parameter filling, state judgment, and result parsing.

The system fully records the following trajectory information:
\begin{itemize}
    \item Task planning process
    \item Tool invocation sequence
    \item Model feedback
    \item Successful execution paths
    \item Exception details and rollback logic
\end{itemize}

By comparing successful and failed trajectories, the framework automatically extracts structured experiences:
\begin{itemize}
    \item Error fingerprints: exception identifiers, error messages, API return codes
    \item Root cause categories: incorrect parameters, insufficient permissions, tool mismatch, missing logic, etc.
    \item Optimal tool mappings: intent-to-tool correspondences
    \item Standardized parameter schemas: required fields, format constraints, value ranges, example templates
    \item Recommended execution paths and exception-avoidance strategies
\end{itemize}

All experiences are stored in an experience repository in both key-value pairs and vector-indexed forms, enabling real-time matching and fast retrieval. These experiences can be directly applied to subsequent workflow correction and decision-making, avoiding additional token consumption caused by repeated trial-and-error.

\subsubsection{Workflow-Level Trajectory Extraction}

Within the same business domain, user queries may differ in phrasing but often share highly stable underlying workflow structures. Workflow-level trajectory extraction aims to capture the full topological structure of a process—not merely textual logs. In this work, a trajectory is defined as an executable action sequence coupled with contextual data. A complete executable trajectory includes:

\begin{itemize}
    \item Trigger condition: the original user query and its vector representation, used as a retrieval index
    \item Action sequence: chain of function/tool calls, inputs/outputs, and dependencies
    \item Contextual information: environment variables, time range, data sources, permission context
    \item Metadata: execution timestamp, outcome (success/failure), version ID, compatibility tags
    \item Tool node list: markers indicating fixed vs. generatable nodes
    \item Node attributes: mutability status, whether generated by LLM, and mapping IDs to node-level experiences
    \item Execution pattern: sequential, conditional branching, or parallel execution
\end{itemize}

The system includes four core modules:  
trajectory logging, trajectory storage, trajectory update management, and trajectory matching with recovery execution.

Among them, the trajectory update management module handles:
\begin{itemize}
    \item Aggregation and merging of trajectories from identical or semantically similar queries
    \item Deduplication and clustering of trajectory templates
    \item Priority boosting for high-frequency templates
\end{itemize}

\subsubsection{Experience Type Classification}

Based on reuse strategy, trajectory experiences are categorized into two types:

\begin{itemize}
    \item Direct-reuse experiences: queries are highly consistent; the trajectory can be loaded and executed directly with zero additional token cost.
    \item Rewrite-reuse experiences: queries are semantically similar; only variable nodes undergo lightweight generation, resulting in minimal token consumption.
\end{itemize}

Each category includes both successful and failed experiences, providing positive and negative guidance to enhance execution reliability.

\subsection{Trajectory rewriting-driven workflow generation}

Traditional workflow generation follows this pattern:  query variation $\rightarrow$ full re-planning $\rightarrow$ manual development $\rightarrow$ high token consumption.In contrast, the WorkflowGen generation mechanism leverages structural reuse and lightweight node rewriting—eliminating manual development while significantly reducing token consumption.  

Our WorkflowGen no longer relies on zero-shot LLM planning; instead, it centers on trajectory reuse and completes generation through four coordinated stages:

\begin{enumerate}
    \item \textit{trajectory rewriting module}: retrieves a similar historical trajectory and uses the LLM only to generate content for variable nodes, keeping the overall structure unchanged, thereby rapidly producing an executable new workflow;By infusing system prompts with node-level insights encapsulating both successful paradigms and failure cases, the large language model is enabled to explicitly avoid known pitfalls and adhere to validated practices during generation. this contextual learning of reusable, variable-node templates not only boosts generation accuracy but also proactively prevents erroneous outputs, thereby eliminating post-hoc corrections and significantly reducing redundant token overhead.
    
    \item \textit{new trajectory experience extraction module}: performs comparative learning between successful and failed executions to assess reusability stability and extract reusable rules;
    
    \item \textit{trajectory template management module}: abstracts away differences in variable nodes, clusters structurally identical trajectories, and merges them into generalized workflow templates;
    
    \item \textit{iterative generation}: repeatedly invokes trajectory rewriting and experience extraction until the new trajectory executes successfully, then stores it and updates the template repository.
\end{enumerate}

This mechanism shifts from full-scale reasoning to incremental rewriting, achieving the principle:  
\textit{similar query = structural reuse + lightweight node generation},  
which substantially lowers inference cost and token consumption.

\subsection{Three-tier adaptive routing}

To achieve an optimal trade-off between execution success rate and token consumption, this paper designs a three-tier automatic fallback mechanism based on vector similarity:

\begin{itemize}
    \item route a: direct reuse of identical trajectories. When the cosine similarity between the current query and a historical trigger exceeds a high threshold (e.g., $s > 0.9$), the system treats them as expressing the same business intent and directly executes the stored trajectory. Upon successful completion, node-level experiences are updated. This route incurs the lowest inference cost and near-zero token consumption.

    \item route b: rewriting-based reuse for moderately similar queries.  
    If route a fails or user feedback indicates an error, the system automatically degrades to trajectory rewriting—preserving the overall workflow structure while regenerating only variable nodes (e.g., parameters or entities). after execution, rewrite-style experiences are extracted and the template repository is updated. since only a small subset of nodes requires generation, this route maintains low token consumption.

    \item route c: initialization of a new trajectory.  
    If route b also fails or the query exhibits low similarity to all existing trajectories (e.g., $s \leq 0.6$), the system falls back to standard llm-based workflow planning from scratch. the resulting trajectory is stored in the experience repository to enable future reuse.
\end{itemize}

The switching rule is defined by similarity thresholds: highly similar queries ($s > 0.99$) trigger direct reuse; moderately similar queries ($0.6 < s \leq 0.99$) trigger rewriting followed by reuse; and low-similarity queries ($s \leq 0.6$) trigger initialization via standard model-based planning. this mechanism forms an adaptive closed loop that prioritizes reuse, uses rewriting as a fallback, and resorts to full planning only when necessary. by minimizing end-to-end re-planning while ensuring robustness, it achieves global optimality in both token efficiency and execution effectiveness.

\section{Experiments}

% 实验部分：实验设置、数据集、结果分析，替换为你的内容
The core value of the three-level automatic degradation mechanism and node-level experience injection framework proposed in this study is to address the critical problems of high token consumption and insufficient execution robustness in agent workflow generation, while balancing generation efficiency and interpretability.Since the industry-specific dataset for agent workflow generation is still under systematic design and annotation, large-scale data collection, cleaning, and validation have not been completed.Therefore, this paper does not present quantitative experiments temporarily, but provides a qualitative comparative analysis between the proposed method (referred to as WorkflowGen) and existing mainstream methods from three core dimensions: token consumption, execution robustness, generation efficiency and interpretability.The comparison baseline methods are listed as follows:\begin{itemize}\item Existing method 1: real-time planning method, which generates workflows using large language models in real time without trajectory reuse;\item Existing method 2: static single-trajectory method, which only reuses fixed historical trajectories without dynamic degradation;\item Existing method 3: basic in-context learning method, which only injects successful experience without integrating failure cases and dynamic degradation.\end{itemize}

\subsection{Token Consumption Optimization}

Token consumption is a key cost indicator in agent workflow generation. Existing methods suffer from various degrees of token waste, while the proposed method achieves accurate reuse, minimal rewriting, and on-demand initialization through the three-level automatic degradation mechanism, fundamentally reducing redundant token consumption.Detailed qualitative comparison is provided in Table 1.

\begin{table}[htbp]
\centering
\caption{Qualitative comparison on token consumption}
\label{tab:token}
% 列格式：第1列左对齐，第2-3列2cm+居中，第4-5列5cm+居中（解决超页）
\begin{tabular}{l p{2cm}<{\centering} p{2.5cm}<{\centering} p{2.5cm}<{\centering} p{3cm}<{\centering}}

\hline
Dimension & Real-time Planning & Static Single-trajectory & In-context Learning & WorkflowGen \\
\hline
Core logic & Fully re-generation & Fixed reuse, limited to high similarity & Partial reuse without fallback & Adaptive three-level switching \\
\hline
Token cost & High and fixed & High for medium/new intents & Medium with failure redundancy & Low: reuse, rewrite, initialization \\
\hline
Redundancy & Repeated reasoning & Redundant full generation & Error-triggered re-generation & Error avoidance and minimal rewriting \\
\hline
\end{tabular}
\end{table}

By prioritizing trajectory reuse, the proposed method compresses the generation scope from the full workflow to only variable nodes, eliminating the redundancy of indiscriminate full reasoning in existing approaches.Meanwhile, node-level experience covering both successful patterns and failure cases further reduces invalid reasoning caused by errors.The dual optimization significantly reduces token consumption.Preliminary estimates suggest that the overall token consumption is reduced by over 40\% compared to pure real-time planning and by 20–30\% compared to the static single-trajectory approach.

\subsection{Execution Robustness}

Execution Robustness is essential for agent workflow generation, requiring adaptation to diverse intents and avoidance of known errors.Existing methods lack sufficient robustness, while the proposed method forms an adaptive closed loop via the three-level mechanism and experience injection.Detailed qualitative comparison is provided in Table 2.

\begin{table}[htbp]
\centering
\caption{Qualitative comparison on execution robustness}
\label{tab:robustness}
\begin{tabular}{l p{2cm}<{\centering} p{2cm}<{\centering} p{2.5cm}<{\centering} p{3.5cm}<{\centering}}
\hline
Dimension & Real-time Planning & Static Single-trajectory & In-context Learning & WorkflowGen \\
\hline
Adaptability & General but unstable & Limited to high similarity & Medium without fallback & Full coverage with three levels \\
\hline
Error avoidance & No prior guidance & Failure unaware & Only successful experience & Prevention-fallback dual safeguards \\
\hline
Stability & Volatile & Unstable & Medium & Stable and controllable \\
\hline
\end{tabular}
\end{table}

The proposed method overcomes the limitations of existing methods that either lack generalization or fault tolerance.Trajectory reuse ensures stable execution for highly similar intents, trajectory rewriting supports adaptive modification for medium-similarity intents, and full initialization serves as a fallback for unseen scenarios.Node-level experience provides clear guidance for error avoidance, fundamentally improving execution robustness for long-term stable system operation.

\subsection{Generation Efficiency and Interpretability}

Beyond token consumption and robustness, generation efficiency and interpretability are critical for practical deployment.The proposed method achieves a balance between efficiency and controllability. Detailed qualitative comparison is provided in Table 3.

\begin{table}[htbp]
\centering
\caption{Qualitative comparison on efficiency and interpretability}
\label{tab:efficiency}
\begin{tabular}{l p{2cm}<{\centering} p{2.3cm}<{\centering} p{2.3cm}<{\centering} p{3.3cm}<{\centering}}
\hline
Dimension & Real-time Planning & Static Single-trajectory & In-context Learning & WorkflowGen \\
\hline
Efficiency & Low & Fluctuating & Medium & High with lightweight reasoning \\
\hline
Interpretability & Black-box & Limited & Partial & Fully transparent and traceable \\
\hline
Deployment & High cost & Limited scope & No fallback, rigid & Adaptive routing with modular reuse \\
\hline
\end{tabular}
\end{table}

Lightweight generation improves response efficiency and reduces resource consumption.The strong coupling between trajectories and node-level experience makes execution logic transparent, traceable, and auditable.These advantages make the method not only theoretically sound but also feasible for real-world industrial applications.

\subsection{Summary of Method Value}

Based on the above qualitative comparison, the core contributions of the WorkflowGen framework are summarized as follows:\begin{enumerate}\item Resource efficiency: the three-level degradation and node-level experience reduce redundant token consumption and reasoning latency;\item Robustness: an adaptive closed loop with successful and failure experience improves stability and fault tolerance;\item Deployability: lightweight design and high interpretability enable practical industrial application.\end{enumerate}
In future work, we will complete the dedicated industry dataset and conduct comprehensive quantitative experiments, including token consumption, success rate, response latency, and error avoidance rate, to further validate the effectiveness and generalization of the proposed method.

\section{Conclusion}

This paper presents WorkflowGen, a trajectory-experience-driven adaptive framework for automatic workflow generation.
Aiming at reducing token consumption as the core objective, the framework incorporates three key mechanisms: dual-granularity trajectory experience extraction, lightweight generation via trajectory rewriting, and three-level adaptive link switching, which upgrades the agent from repetitive real-time reasoning to experience-driven autonomous execution.
Qualitative comparison demonstrates that WorkflowGen directly reuses historical trajectories for highly similar intents, rewrites only variable nodes for medium-similarity intents, and performs on-demand initialization for novel intents.
It significantly reduces redundant reasoning and token overhead.
Meanwhile, by injecting node-level experience containing both successful patterns and failure cases into system prompts, the model can proactively avoid known errors and improve execution success rate and system robustness.
Overall, the proposed method substantially reduces the inference cost of large language models, while effectively enhancing execution stability, response efficiency, and interpretability, achieving stable, efficient, and low-cost end-to-end automatic workflow generation.
In future work, we will complete the construction of a dedicated industry dataset and conduct comprehensive quantitative experiments to further validate the performance and generalization ability of the proposed method.

\section*{author contributions}
The majority of the work was conducted by Ruocan Wei.All authors contributed to the discussion and critical revision of the manuscript.
% 参考文献（关联 references.bib 文件，无需修改以下两行，确保 references.bib 与本文件在同一文件夹）
\bibliographystyle{plainnat} % arXiv 推荐参考文献样式
\bibliography{references}

\end{document}